\documentclass[10pt,twocolumn,letterpaper]{article}

\usepackage[pagenumbers]{wacv} % To force page numbers, e.g. for an arXiv version

\usepackage{graphicx}
\usepackage{amsmath}
\usepackage{amssymb}
\usepackage{booktabs}
\usepackage{float}

\usepackage[pagebackref,breaklinks,colorlinks]{hyperref}
\usepackage{amsmath,amsfonts,bm}

\def\eqref#1{equation~\ref{#1}}
\def\1{\bm{1}}

\DeclareMathAlphabet{\mathsfit}{\encodingdefault}{\sfdefault}{m}{sl}
\SetMathAlphabet{\mathsfit}{bold}{\encodingdefault}{\sfdefault}{bx}{n}

\usepackage[capitalize]{cleveref}
\crefname{section}{Sec.}{Secs.}
\Crefname{section}{Section}{Sections}
\Crefname{table}{Table}{Tables}
\crefname{table}{Tab.}{Tabs.}

\def\wacvPaperID{563} % *** Enter the WACV Paper ID here
\def\confName{WACV}
\def\confYear{2025}

\begin{document}

%%%%%%%%% TITLE - PLEASE UPDATE
\title{A Data Perspective on Enhanced Identity Preservation for Diffusion Personalization}

% \author{First Author\\
% Institution1\\
% Institution1 address\\
% {\tt\small firstauthor@i1.org}
% % For a paper whose authors are all at the same institution,
% % omit the following lines up until the closing ``}''.
% % Additional authors and addresses can be added with ``\and'',
% % just like the second author.
% % To save space, use either the email address or home page, not both
% \and
% Second Author\\
% Institution2\\
% First line of institution2 address\\
% {\tt\small secondauthor@i2.org}
% }

\author{Xingzhe He\textsuperscript{1}\footnote{work done during internship at Adobe} \hspace{5mm}
Zhiwen Cao\textsuperscript{2} \hspace{5mm}
Nicholas Kolkin\textsuperscript{2} \hspace{5mm}
Lantao Yu\textsuperscript{2} \hspace{5mm}
Kun Wan\textsuperscript{2} \\
Helge Rhodin\textsuperscript{1} \hspace{5mm}
Ratheesh Kalarot\textsuperscript{2}\\
{University of British Columbia\textsuperscript{1}, Adobe\textsuperscript{2}}\\
{\tt\small \{xingzhe,rhodin\}@cs.ubc.ca, \{zhiwenc,kolkin,lantaoy,kuwan,kalarot\}@adobe.com}
}

\maketitle

\begin{abstract}
Large text-to-image models have revolutionized the ability to generate imagery using natural language. However, particularly unique or personal visual concepts, such as pets and furniture, will not be captured by the original model. This has led to interest in how to personalize a text-to-image model. Despite significant progress, this task remains a formidable challenge, particularly in preserving the subject's identity. Most researchers attempt to address this issue by modifying model architectures. These methods are capable of keeping the subject structure and color but fail to preserve identity details. Towards this issue, our approach takes a data-centric perspective. We introduce a novel regularization dataset generation strategy on both the text and image level. This strategy enables the model to preserve fine details of the desired subjects, such as text and logos. Our method is architecture-agnostic and can be flexibly applied on various text-to-image models. We show on established benchmarks that our data-centric approach forms the new state of the art in terms of identity preservation and text alignment.
\end{abstract}
\section{Introduction}

\begin{figure*}
\begin{center}
  \includegraphics[width=0.98\textwidth]{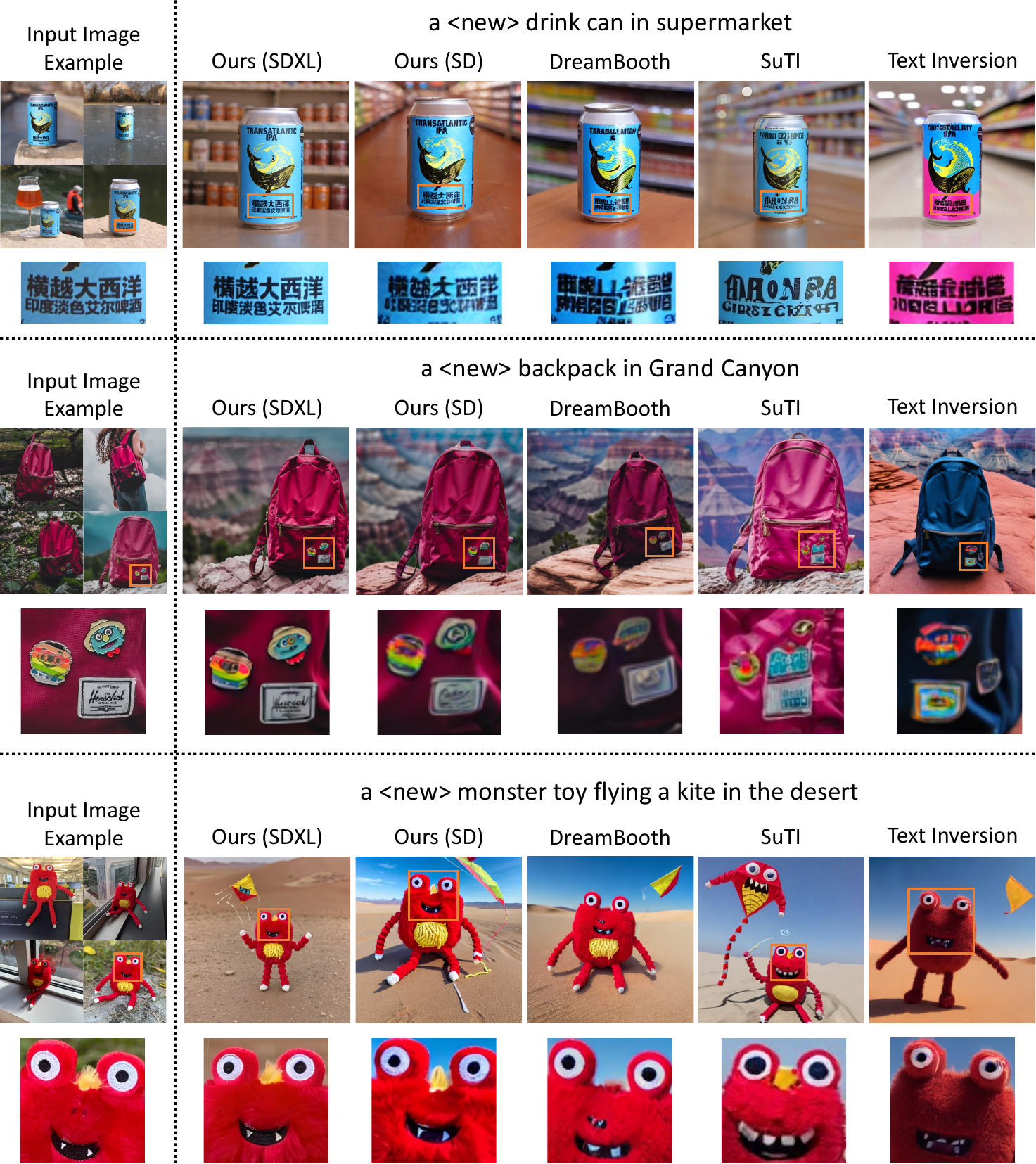}
\end{center}
   \caption{\small Demonstration of our method applied to both SDXL and SD, compared with several prior works. Note that our results preserves very fine details of the subject without overfitting to the training images and losing text alignment.}
\label{fig:teaser}
\end{figure*}

Recent advancements in text-to-image models have greatly improved the creation of diverse and high-quality images based on text input~\cite{rombach2022high, saharia2022photorealistic, podell2023sdxl}. However, these models still face challenges in accurately generating images of specific subjects, such as unique backpack designs or individual dog breeds. Text, being a vague representation, cannot fully capture the intricate details of real-world subjects. This limits the models' capability in generating novel images of the exact subjects.
\looseness=-1

To tackle this issue, previous research has explored fine-tuning large-scale diffusion models, such as DreamBooth~\cite{ruiz2023dreambooth} and TextInversion~\cite{gal2022image}, to associate an identifier token with the desired subject. This not only preserves subject identity but also exhibits impressive generalization across various text inputs~\cite{ruiz2023dreambooth, gal2022image}. Nevertheless, these methods still face challenges in reproducing fine details of the subject, as shown in Figure~\ref{fig:teaser}. \looseness=-1 

We argue the reason that the previous methods fail to preserve identity details, e.g. text and logos, as shown in Figure~\ref{fig:teaser}, Figure~\ref{fig:more_comp1}, \ref{fig:more_comp2}, \ref{fig:more_comp3} in Appendix~\ref{sec:more_results}, is because they cannot capture enough training information~\cite{hu2021lora, kumari2023multi, tewel2023key, qiu2023controlling, han2023svdiff, voynov2023p, sohn2023styledrop}. 
% DreamBooth~\cite{ruiz2023dreambooth} proposed to 1) use an identifier in the text input ``a [identifier token] [class noun]"" to describe all training images, and 2) utilize a regularization dataset, where images are generated by the text input ``a [class noun]"", to prevent overfitting. However, using the simplified text input ``a [identifier token] [class noun]"" may associate unrelated subjects, such as background, instead of the desired subject, to the identifier. Thus, the overfitting issue still exists.

A naive way to capture more training information is to fine-tune the model with more iterations. However, a longer training period can lead to overfitting. As a result, the model ignores text inputs and can only generate images that are visually similar to training images.
% We introduce a novel technique for creating a regularization dataset tailored to counteract overfitting. Fundamental to our method is the observation that the fine-tuning not only associates the subject to the identifier token but also to the same simple sentence structure used for fine-tuning, limiting the ability to use more complex prompts at test time, as shown in Figure~\ref{fig:overfit_comparison} and Figure~\ref{fig:rephrase} \lantao{This sentence is too long. }.
To address the overfitting issue, DreamBooth~\cite{ruiz2023dreambooth} proposed a regularization dataset where images are generated by the text input ``a [class noun]'' (e.g., ``a backpack''). The regularization dataset is used to fine-tune the model, along with the training images, which are prompted as ``a [identifier\_token] [class noun]'' (e.g., ``a sks backpack''). However, experiment results show that the overfitting issues still exist (Figure~\ref{fig:overfit_comparison}). As a result, their method has to stop the fine-tuning early and cannot preserve the fine details.

We propose a new regularization dataset generation strategy to solve the over-fitting issue. We argue that the root cause of ineffectiveness of DreamBooth lies in their over-simplified text input. The simplistic text input ``a [identifier\_token] [class noun]'' and ``a [class noun]'' cannot fully describe the image information. Thus, the identifier token of the desired subject fails to distinguish between the foreground and background, and associate with the entire image.

% \zhiwen{We propose a new regularization dataset generation strategy to solve the over-fitting issue. We argue that the root cause of ineffectiveness of DreamBooth's regularization dataset lies in their over-simplified text input. The simple text input ``a [class noun]"" cannot fully describe the image information. Thus, the identifier token of the desired subject fails to distinguish between the foreground and background, leading to its association with the entire image.}

% \zhiwen{delete this: \textit{Inspired by DreamBooth~\cite{ruiz2023dreambooth}, we propose to generate a regularization dataset to prevent overfitting. However, we do not use the over-simplified text input ``a [identifier\_token] [class noun]"" for all training images. We argue that the simplified text input cannot fully describe the image information and thus the identifier may be also associated to background.}} 
We found that introducing the background text for training examples, ``a [identifier\_token] [class noun] [background]'', and generating regularization dataset with the same background text, effectively associate the identifier token with the desired subject. This approach does not associate the identifier with the entire image, and as a result, it prevents the model from overfitting. The text of the regularization dataset are generated in the format ``a [shape] [color] [texture] [class noun] [background]'' by Large Language Models (LLMs), and the corresponding images are generated from pre-trained diffusion models.
% This enables us to automatically create a variety of prompts with different content and slightly varied sentence structures. These prompts are then used to generate diverse images depicting the subject class in various contexts.

% Therefore, to ensure the proper association of the identifier token with the subject, it's crucial to avoid the model linking the subject to the entire input text sentence. We achieve this by using a combination of Large Language Models (LLMs) and text-to-image diffusion models. 

Our new regularization dataset generation strategy yields a more stable fine-tuning process that can continue longer without overfitting and significantly enhances both identity preservation and text alignment. As demonstrated in Figure~\ref{fig:teaser}, our approach excels in preserving subject-specific details while maintaining robust generalization across diverse text contexts. \looseness=-1

This paper validates the effectiveness of our approach through extensive experimentation on the DreamBench dataset~\cite{ruiz2023dreambooth}, which contains a broad range of subjects, including complex items like drink cans and backpacks, as well as living animals with diverse poses and movements. Our experiments encompass a range of tasks, including subject re-contextualization, attribute modification, accessorization, and style transfer. Our results confirm our method's ability to proficiently preserve subject identity, including intricate details like logos, while demonstrating robust generalization across diverse text inputs. Additionally, we highlight limitations in the original evaluation metric CLIP-T score used in DreamBooth~\cite{ruiz2023dreambooth} and propose an enhanced version of CLIP-T tailored to offer more precise and informative indications of text-image alignment.

We summarize our contributions in the following aspects:\\
1. A data-oriented approach to tackle overfitting issues during diffusion model fine-tuning.\\
2. A modification to the CLIP-T score to enhance its capacity for indicating text alignment.\\
3. A systematic exploration of diverse methods for generating a regularization dataset, with the intention of providing valuable insights to the broader research community.

\section{Related Work}

% Text-conditioned image editing methods~\cite{brooks2023instructpix2pix, kawar2023imagic} enable image manipulation based on text input while retaining the subject's identity. However, these methods struggle when tasked with generating entirely new images in different contexts. On the other hand, image composition techniques~\cite{wu2019gp, lin2018st, cong2020dovenet, yang2023paint} can transplant a given subject into alternative backgrounds but cannot create the subject within a novel scene from text alone.

\textbf{Diffusion Based Text-to-Image Models} have recently demonstrated remarkable progress, primarily through the utilization of diffusion models~\cite{ho2020denoising}. These achievements are particularly pronounced when employing large models trained on extensive datasets~\cite{rombach2022high, podell2023sdxl, saharia2022photorealistic, balaji2022ediffi, nichol2021glide, ramesh2022hierarchical}. Our method extends the pre-trained diffusion models to incorporate personalized concepts. We apply this method to well-established models, such as StableDiffusion~\cite{rombach2022high} and StableDiffusionXL~\cite{podell2023sdxl}. Note that our method is fundamentally rooted in data, and is potentially applicable to a wide range of different diffusion model architectures.

\textbf{Text-to-Image Personalization} aims to imbue pre-trained diffusion models with the ability to produce novel images of specific subjects~\cite{ruiz2023dreambooth, gal2022image}. However, previous methods grappled with a delicate balancing act. They oscillate between underfitting the intended subject and overfitting to the training dataset and losing text alignment.
\cite{hao2023vico} attempt to address the problem of underfitting the subject by injecting a reference image feature map into attention modules to preserve identity.~\cite{chen2023disenbooth}, on the other hand, embark on the task of disentangling subjects from backgrounds by employing a CLIP image encoder for encoding backgrounds~\cite{radford2021learning}.
To combat the issue of overfitting, researchers have explored diverse strategies, mostly by fine-tuning different decomposition parts of the model~\cite{hu2021lora, kumari2023multi, tewel2023key, qiu2023controlling, han2023svdiff, voynov2023p, sohn2023styledrop}.
To sidestep time-consuming fine-tuning the diffusion model, %Confronting the protracted fine-tuning times,
several encoder-based methods have emerged, involving the training of an additional encoder for all subjects~\cite{shi2023instantbooth, gal2023designing, ruiz2023hyperdreambooth, arar2023domain, li2023blip}. These encoder-based methods often yield increased diversity in generated content, albeit with some trade-offs in identity preservation compared to fine-tuning-based approaches.
Our approach focuses on maximizing quality at the cost of training time. It improves both identity preservation and text alignment relative to prior work. We leave the topic of accelerated training as an avenue for future exploration.

\textbf{Involving More Data} is our core idea. Previous work~\cite{wang2023reprompt} has shown that a better prompt can significantly improve the generated image quality. The most similar approach to ours is proposed in DreamBooth~\cite{ruiz2023dreambooth}. They propose to generate a regularization dataset using prompt ``a [class noun]'', e.g., ``a backpack'', to alleviate overfitting. However, our experiments reveal that this generated dataset lacks the strength to effectively counter overfitting.~\cite{kumari2023multi} extract the real text-image pairs that similar to the training examples from existing dataset, but they still need to restrict the number of training iterations to mitigate overfitting risks.~\cite{chen2023subject} speed up customization with a feed forward network trained on many datasets of custom objects, but they prioritize speed over quality improvements.~\cite{ma2023subject} propose to generate masks, bounding boxes, and their corresponding tags to train a better adapter and thus a better composition method. Our work is primarily centered on enhancing generation quality, with the aspect of speed left for future exploration.
\section{Method} \label{sec:method}

Our main goal is to fine-tune a pre-trained text-to-image diffusion model using a limited set of subject-specific training examples, typically around 4 to 6 images. This fine-tuning process is aimed at enabling the diffusion model to generate new images of the provided specific subject. %However, preventing the model from overfitting to the training examples is a challenging task. 

\begin{figure*}[t]
\begin{center}
  \includegraphics[width=\textwidth]{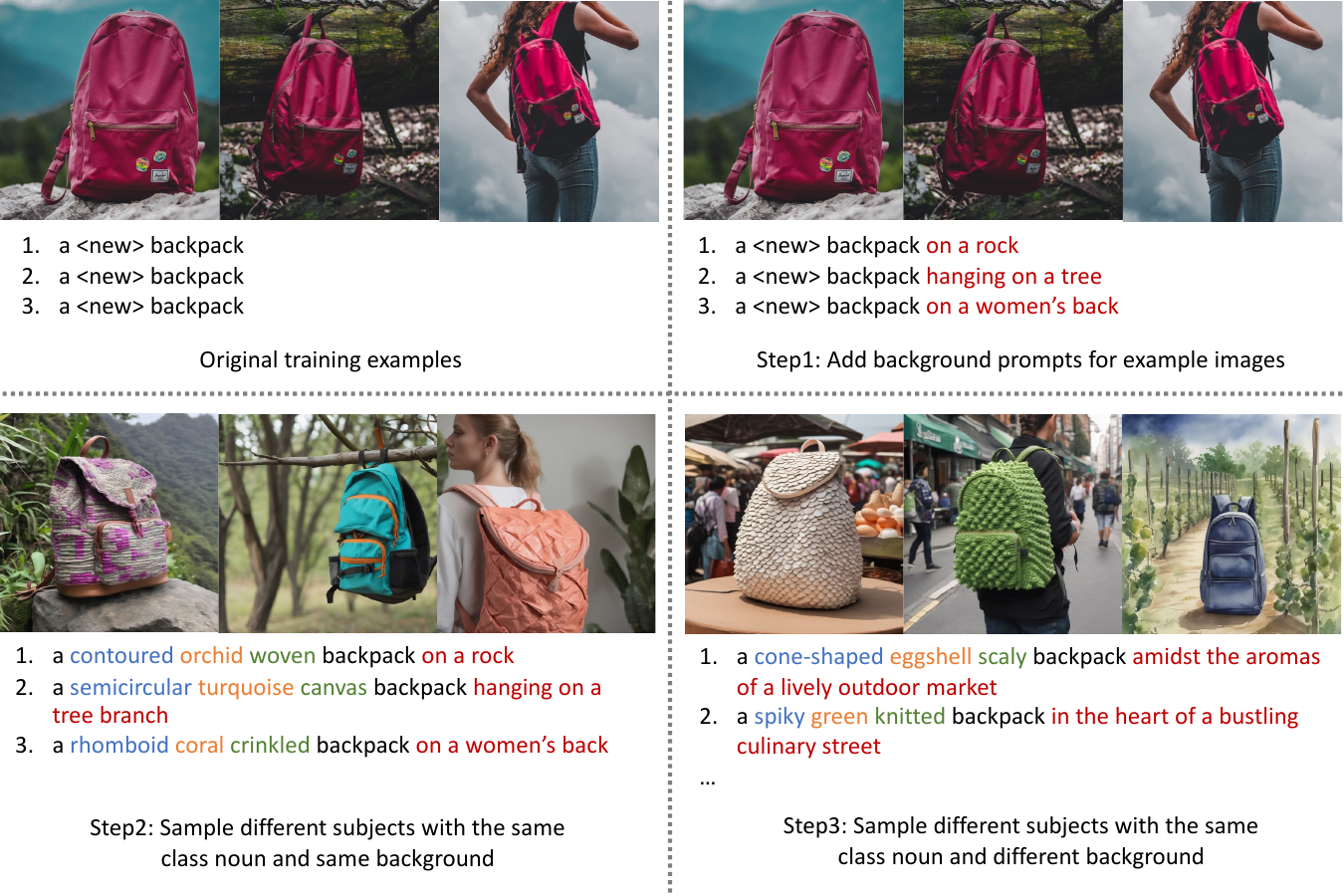}
\end{center}
   \caption{\small \textbf{Overview.} In contrast to prior approaches, we introduce specific prompts to the training examples and create a regularization dataset with a wider range of images guided by these prompts. To further boost diversity of regularization dataset, we generate additional prompts and images using the same prompt formats.}
\label{fig:overview}
\end{figure*}

\textbf{Recap of DreamBooth}.
Instead of linking the new subject with a noun word~\cite{gal2022image}, DreamBooth~\cite{ruiz2023dreambooth} proposed to linking the new subject with an adjective word, by inserting a special identifier token before the subject class noun. For example, ``a backpack'' becomes ``a [identifier\_token] backpack'' or ``a \textit{sks} backpack'' in DreamBooth \cite{ruiz2023dreambooth}. Furthermore, to prevent overfitting, they proposed to create a dataset of text-image pairs by employing prompts in the format ``a [class noun]'' (e.g., ``a backpack''). This dataset serves as a counterbalance to the training examples. During training, both the training examples and the regularization examples are fed into the diffusion model. 
However, we discovered that the divergence of the over-simplified text and the image still leads to overfitting before fine details are fully learned.

\textbf{Our Approach}. 
In contrast to previous methods using a simplistic prompt for the real training examples, we first introduced individual and more concrete prompts, such as ``a backpack on a rock''. Surprisingly, this adjustment did not yield significant improvements, and the model still exhibited rapid overfitting. It extends the fine-tuning time only when combined with a strong regularization dataset using prompts that describe foreground and background in more details. Inspired by this observation, our focus shifted towards automatically constructing an appropriate regularization dataset to address the issue of overfitting.

The overview of our method is presented in Figure~\ref{fig:overview}. Importantly, our approach retains the model architecture while concentrating on the development of a novel regularization dataset to combat overfitting effectively. This means that our method can be combined with different pre-trained diffusion models.

\textbf{Enhancing Training with Prompts}. 
The inclusion of prompts in training instances plays a pivotal role in our approach. While employing a uniform prompt format such as ``a [identifier\_token] [class noun]'' for all training examples may seem straightforward, it inadvertently increases the risk of associating the token with both foreground and background, causing overfitting to the training data. To address this challenge, we developed this idea further by incorporating background descriptions into each image, as depicted in the upper-right corner of Figure~\ref{fig:overview}. Additionally, we replaced the provided class name with more specific ones, which will be discussed further in Appendix~\ref{sec:better_naming}. It's important to note that merely adding the background prompt alone does not yield the desired benefits. Instead, our contribution is on constructing a dedicated regularization dataset, which proves to be essential in addressing the issue of overfitting, as elaborated in subsequent sections.

\textbf{Generating regularization examples against training prompts}. 
We present a novel approach for generating a regularization dataset, which overcomes the ineffectiveness of the original example prompt. For instance, let's consider the initial prompt ``a backpack on a rock''. We enhance this prompt by introducing structural components in the format of ``a [shape] [color] [texture] [class noun] [background]'', where [shape], [color], and [texture] represent randomly selected adjectives drawn from a pool of 100 options for each attribute. These adjectives are then combined to create prompts, such as ``a contoured orchid woven backpack on a rock'' and ``a semicircular turquoise canvas backpack hanging on a tree branch''. The created prompts are used for image generation, as demonstrated in the lower-left corner of Figure~\ref{fig:overview}. The pools of the adjectives are generated automatically by Large Language Models~\cite{chatgpt}. For more detailed information on this process, please refer to Appendix~\ref{sec:formatted_prompt_generation}.

\textbf{Amplifying Diversity with Structured Prompts}. 
To increase the diversity of our regularization dataset, we randomly generate 500 additional background phrases and 100 style phrases. We then combine these phrases in a random manner, resulting in structured prompts in the format ``a [style] [shape] [color] [texture] [class noun] [background]''. For instance, this approach generates prompts like ``a photo of a cone-shaped eggshell scaly backpack amidst the aromas of a lively outdoor market'' and ``a children's storybook illustration of a trapezoidal coral embossed backpack against the canvas of a city skyline'', as showcased in the lower-right corner of Figure~\ref{fig:overview}. 
%We randomly exclude the style prompt "a photo of".
This strategy substantially enhances image diversity compared to the simplistic use of the prompt ``a [class noun]''. \looseness=-1

\textbf{Adaption for Living Entities}.
For living entities, we use descriptors like body, skin/fur, and emotion instead of shape, color, and texture. Additionally, we introduce motion into the backgrounds, transitioning from static scenes like ``in an urban city'' to dynamic contexts such as ``walking in an urban city''. For a detailed explanation of this adaptation process, please refer to Appendix~\ref{sec:formatted_prompt_generation}.

\textbf{Dropout}.
In our strategy to enrich the dataset, we include three descriptive words for each object, which increases diversity. However, this approach can lead to a potential issue: overfitting to sentence structure. The model might learn that one adjective word corresponds to a specific subject, while three adjective words signify a different subject. To mitigate this, we introduce randomness by randomly excluding some words during training. For example, we may have a prompt like ``a [shape] [color] backpack on a rock'' excluding a texture descriptor. This dropout technique adds an element of unpredictability, strengthening the model's adaptability and reducing its reliance on fixed sentence formats.

\textbf{Cropping}. 
\cite{tewel2023key} point out that the model is prone to overfit to the image layout when attempting to learn personalized concepts from a limited set of examples. Similarly,~\cite{kumari2023multi} observe that employing random cropping enhances convergence speed and yields improved results. To enhance our model's performance, we incorporate random cropping with a variable ratio ranging from 0.75 to 1. It is worth noting that for SDXL, we incorporate cropping coordinates as described by~\cite{podell2023sdxl}, where we increase the original image size to achieve a cropped image size of 1024x1024 pixels.

\textbf{Implementation Details}
We create 2000 regularization images for each subject, distributed as follows: 
20\% photorealistic images have different subjects in same background; 60\% photorealistic images have different subjects in different background; and 20\%  differently styled images have different subjects in different background.
Additional details regarding the optimal iteration range can be found in Appendix~\ref{sec:implementation_details}. \looseness=-1
\section{Experiments}

We quantitatively evaluate on the DreamBench dataset~\cite{ruiz2023dreambooth}, which contains 30 subjects ranging from backpacks and stuffed animals to dogs and cats. The dataset encompasses 25 distinct testing prompts for inanimate objects and 25 for living entities. Following the original paper, we generate four images per prompt, which yields a total of 3000 images for evaluation. Importantly, the testing prompts used in our experimental results are included in our regularization dataset.

Our baseline model is DreamBooth (SD backbone)~\cite{ruiz2023dreambooth}. For a fair comparison, we take the following steps: 1) Employ SDXL to generate 2000 high-quality regularization images; 2) Use ``olis'' as the identifier word; 3) Set the learning rate to 2e-6 for 1500 iterations; and 4) Apply 200 steps of DDIM for inference.

\subsection{Result Comparison} \label{sec:quantitative_comparison}

The qualitative comparison is depicted in Figure~\ref{fig:teaser} and Figure~\ref{fig:more1}-\ref{fig:more_comp3} in Appendix~\ref{sec:more_results}. We use ``new'' to represent the `` [identifier\_token]''. Our approach outperforms alternative methods in preserving intricate subject-specific details, such as logos on drink cans, unique icons on backpacks, and fine hair textures on toy monsters. %Other methods struggle with preserving these details, emphasizing the superior subject fidelity of our approach.

% \begin{figure}
% % \vspace{-4ex}
% \begin{center}
%   \includegraphics[width=0.98\textwidth]{images/more_comp.pdf}
% \end{center}
% \vspace{-2ex}
%    \caption{\small \textbf{Qualitative Comparison}. compared with prior works, our results preserves finer details of the subject without overfitting to the training images or losing text alignment.
% }
% \label{fig:more_comp}
% \end{figure} 

% \begin{figure*}
% \begin{center}
%   \includegraphics[width=0.9\textwidth]{images/more4.pdf}
% \end{center}
%    \caption{\small \textbf{Qualitative Results}.Our results preserves finer details of the subject without overfitting to the training images or losing text alignment.
%     }
% \label{fig:more4}
% \end{figure*}

% \vspace{-4ex}
\begin{table}
\centering
\small
\resizebox{0.98\linewidth}{!}{%
\begin{tabular}{@{\;}llcccc@{\;}}
\toprule
Methods & Backbone & DINO $\uparrow$ & CLIP-I $\uparrow$ & CLIP-T (vague class) $\uparrow$ & CLIP-T (subject) $\uparrow$\\
\midrule
Real Image (on training set) & - & 0.774 & 0.885 & 0.298 & 0.311\\
\midrule
Textual Inversion~\cite{gal2022image} & SD & 0.569 & 0.780 & 0.255 & - \\
Textual Inversion (our impl) & SD & 0.611 & 0.772 & 0.267 & 0.289 \\
DreamBooth~\cite{ruiz2023dreambooth} & SD & 0.668 & 0.803 & \textbf{0.305} & - \\
DreamBooth (our impl) & SD & 0.682 & 0.808 & 0.301 & 0.306 \\
Ours & SD & 0.704 & 0.824 & 0.293 & 0.309 \\
Ours & SDXL & \textbf{0.744} & \textbf{0.842} & 0.297 & \textbf{0.312} \\
\bottomrule
\end{tabular}
}
% \vspace{1ex}
\caption{\small
Evaluation on the DreamBench.
}
% \vspace{-4ex}
\label{tab:clip_matrics}
\end{table}
% \vspace{-10ex}
\begin{table}
\centering
\small
\resizebox{0.98\linewidth}{!}{%
\begin{tabular}{@{\;}lcc@{\;}}
\toprule
 & Ours (SD) vs DreamBooth & Ours (SDXL) vs Ours (SD)\\
\midrule
Subject Alignment & 64.9\% / 35.1\% & 67.8\% / 32.2\% \\
Textual Alignment & 52.8\% / 47.2\% & 46.8\% / 53.2\%\\
\bottomrule
\end{tabular}
}
% \vspace{1ex}
\caption{
Human preference comparison
}
\vspace{-4ex}
\label{tab:human_evaluation}
\end{table}

% \begin{figure}[ht]
%     \begin{minipage}{0.4\textwidth} % Adjust the width as needed
%         \caption{\small \textbf{CLIP-T score on different object names}. We randomly select an image with the prompt ``a teddy bear toy sitting beside a river". Subsequently, we evaluate prompts ``a {} in the grass next to a tree" and ``a {} sitting on a sofa". When employing the vague prompt ``toy", both the ground truth and the mismatched example are classified as mismatches (\textless0.3). In contrast, when using specific class names, the ground truth is categorized as a match (\textgreater0.3), while the mismatched examples remain classified as mismatches.}
%     \end{minipage}\hfill
%     \begin{minipage}{0.55\textwidth} % Adjust the width as needed
%         \includegraphics[width=\linewidth]{images/clip_metic.pdf} % Replace with your image
%     \end{minipage}
%     \label{fig:clip_metric}
% \end{figure}

\begin{figure*}
\begin{center}
  \includegraphics[width=\textwidth]{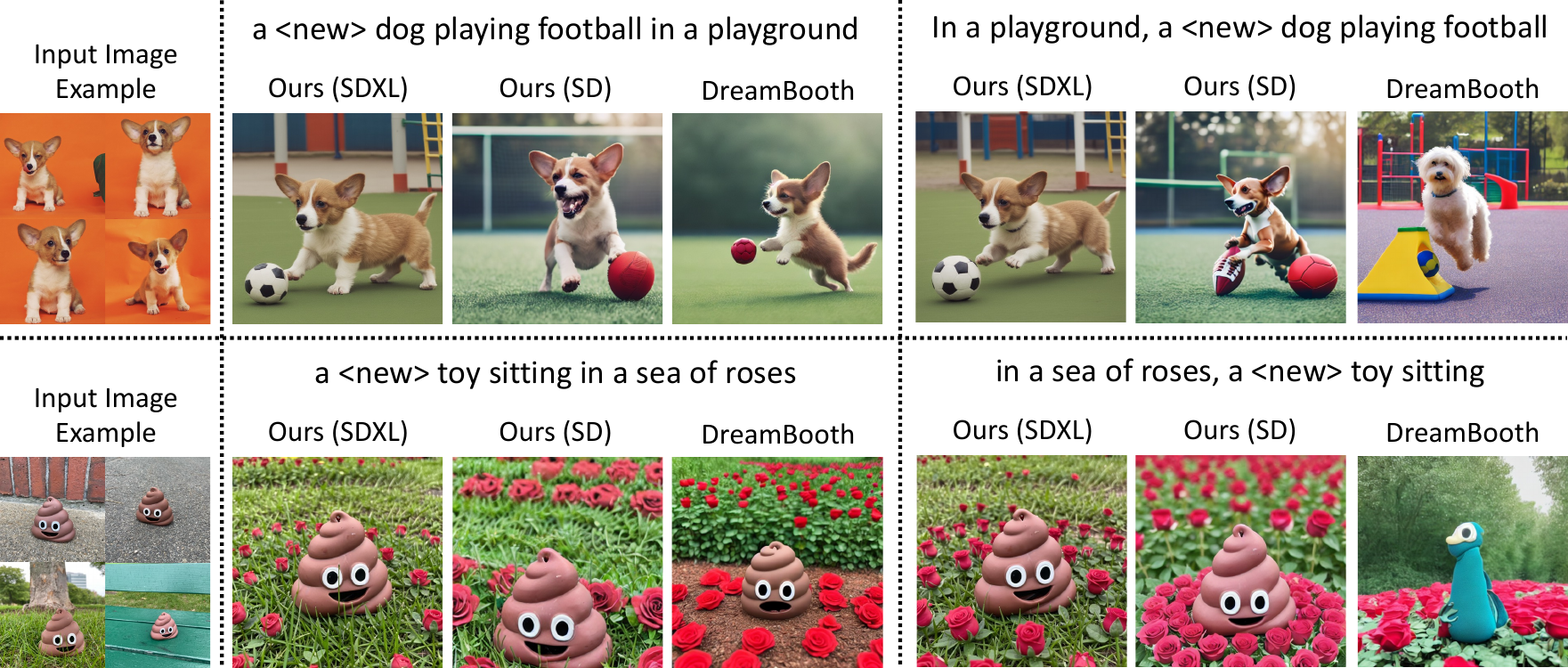}
\end{center}
    \vspace{-0.3cm}
   \caption{\textbf{Rephrasing prompts}. Even when using rephrased prompts, our method maintains subject identity, a quality DreamBooth lacks. Notably, SDXL consistently generates images with minimal deviation from the originals.}
\label{fig:rephrase}
\end{figure*}

\begin{figure*}[t]
\begin{center}
  \includegraphics[width=\textwidth]{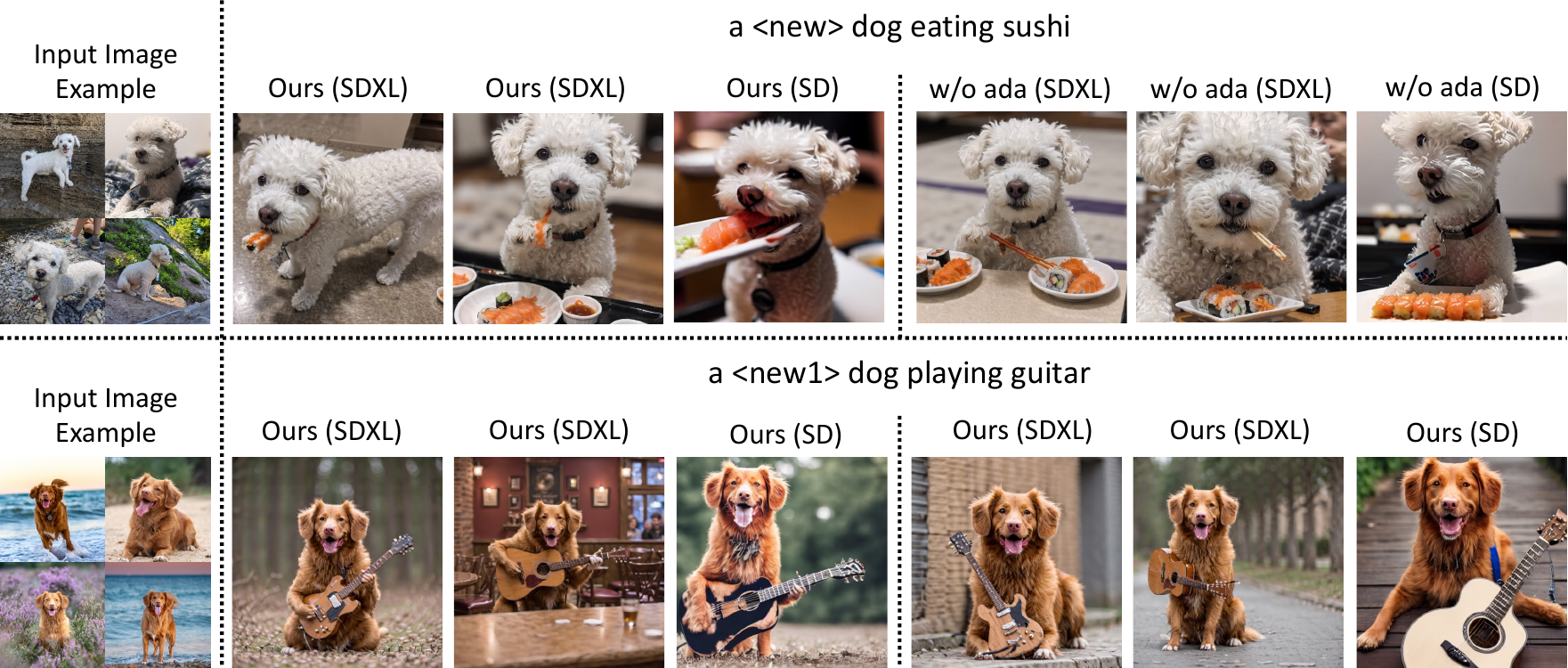}
\end{center}
    \vspace{-0.2cm}
   \caption{\textbf{Influence of adaption to living entities}. Without adaptation, the model might overlook motion in the prompts and focus solely on assembling objects within the images.}
\label{fig:wo_adapt}
\end{figure*}

\begin{figure*}
\begin{center}
% \vspace{-3ex}
  \includegraphics[width=0.7\textwidth]{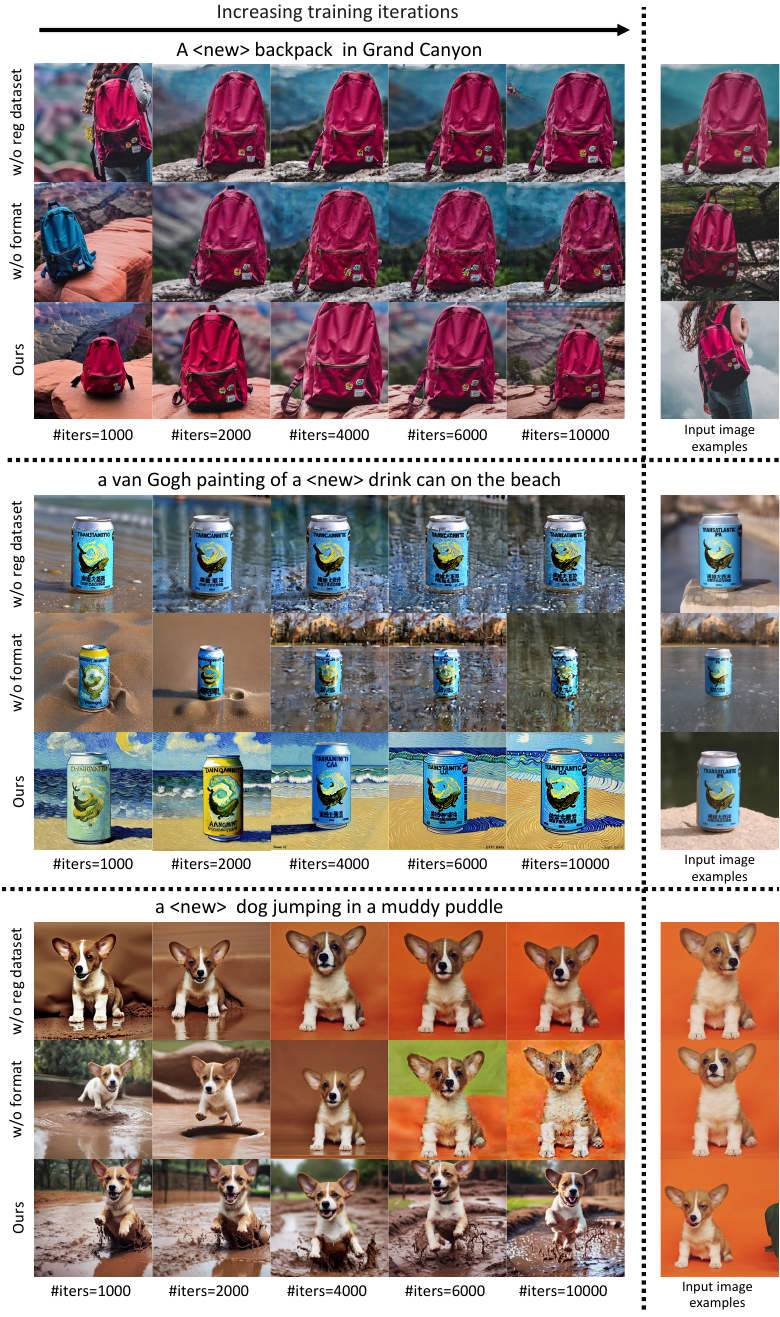}
\end{center}
   \caption{\small \textbf{Overfitting Prevention}. Our regularization dataset effectively prevents the model from overfitting to the training images. Interestingly, when we employ a regularization dataset with a simplistic prompt ``a [class noun]'', the model exhibits improved text alignment during initial iterations, but experiences a decline in performance over time.}
\label{fig:overfit_comparison}
% \vspace{-3ex}
\end{figure*}

In our quantitative assessment, we follow DreamBooth to employ DINO~\cite{caron2021emerging} and CLIP-I~\cite{radford2021learning}, to evaluate subject fidelity. 
For text alignment evaluation, we argue that the CLIP-T used in DreamBooth may yield inaccurate measurements of text-image alignment when vague class names are employed. More details and discussion can be found in Appendix~\ref{sec:better_naming}. To rectify this issue, we replace vague class names with highly specific object names, which results in a substantial improvement in the CLIP-T score for ground truth text-image pairs, as shown in Table~\ref{tab:clip_matrics}.

We conducted a user study to compare with DreamBooth~\cite{ruiz2023dreambooth}. We asked 20 people to answer 30 questions about subject alignment and 30 questions about text alignment, in a total of 1200 questions. The questions are randomly sampled from a large pool of DreamBench test prompts. The details can be found in Appendix~\ref{sec:user_study}. As shown in Table~\ref{tab:human_evaluation}, our method is overwhelmingly preferred for subject alignment and comparable in
text alignment. We also compared the effect of a different base model with SDXL instead of SD. SDXL is more expressive in details and hence leads to better subject alignment. For textural alignment SD works slightly better.

\subsection{Model Analysis}

\textbf{Does the model overfit to the formatted prompts?}
To assess the impact of prompt phrasing variations, we rephrase the prompts and compare them to the originals, as illustrated in Figure~\ref{fig:rephrase}. While our model consistently employs formatted prompts starting with ``a \textless new\textgreater [class noun]'' during training, it demonstrates resilience against overfitting to this specific format. In contrast, DreamBooth displays a notable departure from subject identity. This observation confirms our assertion that the simplistic use of unvaried prompts in the format ``a [class noun]'' for constructing a regularization dataset can lead to overfitting to sentence structure. Additionally, it is noteworthy that SDXL exhibits a higher degree of consistency in generating images in response to rephrased prompts, with minimal deviation from the originals.

\textbf{How important is the formatted prompt?}
We conducted an ablation test with two variations: 1) without using a regularization dataset, and 2) using a regularization dataset with simple prompts in the format ``a [class noun]''. The results are illustrated in Figure~\ref{fig:overfit_comparison}. These findings emphasize the importance of both the regularization dataset and the utilization of well-structured prompts. Notably, when employing a regularization dataset without well-structured prompts, there is an initial increase in diversity during early training phases compared to scenarios without any regularization dataset. However, over extended training iterations, a noticeable decline in model performance becomes evident. We attribute this decline to the incongruity between the overly simplistic prompt format ``a [class noun]'' and the complexities of the generated images. This observation underscores the pivotal role of the well-structured prompts in enhancing model performance, and thus highlights their significance in our approach.

\textbf{Does the adaption to living entities help?}
The key difference between inanimate objects and living things is motion. Training without adaptation tends to overlook motion while still generating objects separately, as shown in Figure~\ref{fig:wo_adapt}. However, it is important to note that adaptation makes generating motion easier, but the model can still produce motion even without it.

% \vspace{-2.5cm}

\textbf{How does the training and regularization dataset size affects the performance?} 
We conducted grid experiments on number of training examples 1, 2, 4, and number of regularization examples 100, 500, 1000, 2000. 
The results are illustrated in Figure ~\ref{fig:ablation} in Appendix~\ref{sec:more_results}. Our method is robust to smaller regularization set size, as only 100 regularization examples can also effectively prevent overfitting and the model is still able to preserve the very fine details of the subjects.
% \zhiwen{Interestingly, when the subject is complex, only one or two training examples with a large amount of regularization examples ($\geq$500) may result in underfitting, such as the backpack in Figure~\ref{fig:ablation} bottom. This is not observed when the subject is simple, such as the dog  in Figure~\ref{fig:ablation} top.}

% \zhiwen{What you want to express above? Simply putting some observation in the paper without further discussion or explanation is not a good idea.}

\textbf{Is the random dropout of adjective words necessary?}
When altering attributes like color, we found that using random dropout is essential. Without it, the model tends to excessively preserve the original identity, and struggle in color changes. With random dropout, the task becomes easier while still preserving identity, as shown in Figure~\ref{fig:color_modification} in Appendix~\ref{sec:more_results}. It's worth noting that for color modification, using ``a \textless new\textgreater [color] [class noun]'' works better than ``a [color] \textless new\textgreater [class noun]''.

\section{Limitations}
Despite its impressive quality, our method has two limitations: 1) It needs to generate a regularization dataset for each category, which increases the overall training time (in practice $<$500 images also work for simple cases); 2) On a single A100 GPU, it requires approximately 1.5 hours for SD and 3.5 hours for SDXL to complete 4000 and 8000 training iterations, respectively.  We leave acceleration to future research.
Currently, we assume that the real images of the target object are annotated manually. In the future, caption generator could be used, such as BLIP~\cite{li2022blip, li2023blip2}.

\section{Conclusion}
We introduced a data-centric approach to enhance diffusion model personalization. We proposed to incorporate formatted prompts and their generated images to generate a structured regularization dataset. Our method effectively reduces overfitting. %, allowing for up to a 5x longer training duration. 
This results in significant improvements in image quality, identity preservation, and diversity of generated samples aligned with input text prompts. Although this approach requires extended training times and increased memory resources, we foresee potential solutions in the future, as our method sets the stage for forthcoming data-centric approaches. Notably, our method focuses on data augmentation rather than model architecture adjustments, making it potentially compatible for different pre-trained text-to-image diffusion models. 
For future work, we would like to focus on 1) generalize to multi-concepts; 2) apply on videos; and 3) accelerate the procedure.

% This approach paves the way for a data-centric paradigm in personalized diffusion algorithms, reducing the reliance on complex model structures.

%%%%%%%%% REFERENCES
{\small
\bibliographystyle{ieee_fullname}
\bibliography{main}

\begin{thebibliography}{10}\itemsep=-1pt

\bibitem{special_token}
2kpr.
\newblock dreambooth-tokens, 2022.

\bibitem{arar2023domain}
Moab Arar, Rinon Gal, Yuval Atzmon, Gal Chechik, Daniel Cohen-Or, Ariel Shamir, and Amit~H Bermano.
\newblock Domain-agnostic tuning-encoder for fast personalization of text-to-image models.
\newblock {\em arXiv preprint arXiv:2307.06925}, 2023.

\bibitem{balaji2022ediffi}
Yogesh Balaji, Seungjun Nah, Xun Huang, Arash Vahdat, Jiaming Song, Karsten Kreis, Miika Aittala, Timo Aila, Samuli Laine, Bryan Catanzaro, et~al.
\newblock ediffi: Text-to-image diffusion models with an ensemble of expert denoisers.
\newblock {\em arXiv preprint arXiv:2211.01324}, 2022.

\bibitem{caron2021emerging}
Mathilde Caron, Hugo Touvron, Ishan Misra, Herv{\'e} J{\'e}gou, Julien Mairal, Piotr Bojanowski, and Armand Joulin.
\newblock Emerging properties in self-supervised vision transformers.
\newblock In {\em Proceedings of the IEEE/CVF international conference on computer vision}, pages 9650--9660, 2021.

\bibitem{chen2023disenbooth}
Hong Chen, Yipeng Zhang, Xin Wang, Xuguang Duan, Yuwei Zhou, and Wenwu Zhu.
\newblock Disenbooth: Disentangled parameter-efficient tuning for subject-driven text-to-image generation.
\newblock {\em arXiv preprint arXiv:2305.03374}, 2023.

\bibitem{chen2023subject}
Wenhu Chen, Hexiang Hu, Yandong Li, Nataniel Rui, Xuhui Jia, Ming-Wei Chang, and William~W Cohen.
\newblock Subject-driven text-to-image generation via apprenticeship learning.
\newblock {\em arXiv preprint arXiv:2304.00186}, 2023.

\bibitem{gal2022image}
Rinon Gal, Yuval Alaluf, Yuval Atzmon, Or Patashnik, Amit~H Bermano, Gal Chechik, and Daniel Cohen-Or.
\newblock An image is worth one word: Personalizing text-to-image generation using textual inversion.
\newblock {\em arXiv preprint arXiv:2208.01618}, 2022.

\bibitem{gal2023designing}
Rinon Gal, Moab Arar, Yuval Atzmon, Amit~H Bermano, Gal Chechik, and Daniel Cohen-Or.
\newblock Designing an encoder for fast personalization of text-to-image models.
\newblock {\em arXiv preprint arXiv:2302.12228}, 2023.

\bibitem{han2023svdiff}
Ligong Han, Yinxiao Li, Han Zhang, Peyman Milanfar, Dimitris Metaxas, and Feng Yang.
\newblock Svdiff: Compact parameter space for diffusion fine-tuning.
\newblock {\em arXiv preprint arXiv:2303.11305}, 2023.

\bibitem{hao2023vico}
Shaozhe Hao, Kai Han, Shihao Zhao, and Kwan-Yee~K Wong.
\newblock Vico: Detail-preserving visual condition for personalized text-to-image generation.
\newblock {\em arXiv preprint arXiv:2306.00971}, 2023.

\bibitem{ho2020denoising}
Jonathan Ho, Ajay Jain, and Pieter Abbeel.
\newblock Denoising diffusion probabilistic models.
\newblock {\em Advances in neural information processing systems}, 33:6840--6851, 2020.

\bibitem{hu2021lora}
Edward~J Hu, Yelong Shen, Phillip Wallis, Zeyuan Allen-Zhu, Yuanzhi Li, Shean Wang, Lu Wang, and Weizhu Chen.
\newblock Lora: Low-rank adaptation of large language models.
\newblock {\em arXiv preprint arXiv:2106.09685}, 2021.

\bibitem{kumari2023multi}
Nupur Kumari, Bingliang Zhang, Richard Zhang, Eli Shechtman, and Jun-Yan Zhu.
\newblock Multi-concept customization of text-to-image diffusion.
\newblock In {\em Proceedings of the IEEE/CVF Conference on Computer Vision and Pattern Recognition}, pages 1931--1941, 2023.

\bibitem{li2023blip}
Dongxu Li, Junnan Li, and Steven~CH Hoi.
\newblock Blip-diffusion: Pre-trained subject representation for controllable text-to-image generation and editing.
\newblock {\em arXiv preprint arXiv:2305.14720}, 2023.

\bibitem{li2023blip2}
Junnan Li, Dongxu Li, Silvio Savarese, and Steven Hoi.
\newblock Blip-2: Bootstrapping language-image pre-training with frozen image encoders and large language models.
\newblock {\em arXiv preprint arXiv:2301.12597}, 2023.

\bibitem{li2022blip}
Junnan Li, Dongxu Li, Caiming Xiong, and Steven Hoi.
\newblock Blip: Bootstrapping language-image pre-training for unified vision-language understanding and generation.
\newblock In {\em International Conference on Machine Learning}, pages 12888--12900. PMLR, 2022.

\bibitem{ma2023subject}
Jian Ma, Junhao Liang, Chen Chen, and Haonan Lu.
\newblock Subject-diffusion: Open domain personalized text-to-image generation without test-time fine-tuning.
\newblock {\em arXiv preprint arXiv:2307.11410}, 2023.

\bibitem{nichol2021glide}
Alex Nichol, Prafulla Dhariwal, Aditya Ramesh, Pranav Shyam, Pamela Mishkin, Bob McGrew, Ilya Sutskever, and Mark Chen.
\newblock Glide: Towards photorealistic image generation and editing with text-guided diffusion models.
\newblock {\em arXiv preprint arXiv:2112.10741}, 2021.

\bibitem{chatgpt}
OpenAI.
\newblock Chatgpt, 2023.

\bibitem{podell2023sdxl}
Dustin Podell, Zion English, Kyle Lacey, Andreas Blattmann, Tim Dockhorn, Jonas M{\"u}ller, Joe Penna, and Robin Rombach.
\newblock Sdxl: Improving latent diffusion models for high-resolution image synthesis.
\newblock {\em arXiv preprint arXiv:2307.01952}, 2023.

\bibitem{qiu2023controlling}
Zeju Qiu, Weiyang Liu, Haiwen Feng, Yuxuan Xue, Yao Feng, Zhen Liu, Dan Zhang, Adrian Weller, and Bernhard Sch{\"o}lkopf.
\newblock Controlling text-to-image diffusion by orthogonal finetuning.
\newblock {\em arXiv preprint arXiv:2306.07280}, 2023.

\bibitem{radford2021learning}
Alec Radford, Jong~Wook Kim, Chris Hallacy, Aditya Ramesh, Gabriel Goh, Sandhini Agarwal, Girish Sastry, Amanda Askell, Pamela Mishkin, Jack Clark, et~al.
\newblock Learning transferable visual models from natural language supervision.
\newblock In {\em International conference on machine learning}, pages 8748--8763. PMLR, 2021.

\bibitem{ramesh2022hierarchical}
Aditya Ramesh, Prafulla Dhariwal, Alex Nichol, Casey Chu, and Mark Chen.
\newblock Hierarchical text-conditional image generation with clip latents.
\newblock {\em arXiv preprint arXiv:2204.06125}, 1(2):3, 2022.

\bibitem{rombach2022high}
Robin Rombach, Andreas Blattmann, Dominik Lorenz, Patrick Esser, and Bj{\"o}rn Ommer.
\newblock High-resolution image synthesis with latent diffusion models.
\newblock In {\em Proceedings of the IEEE/CVF conference on computer vision and pattern recognition}, pages 10684--10695, 2022.

\bibitem{ruiz2023dreambooth}
Nataniel Ruiz, Yuanzhen Li, Varun Jampani, Yael Pritch, Michael Rubinstein, and Kfir Aberman.
\newblock Dreambooth: Fine tuning text-to-image diffusion models for subject-driven generation.
\newblock In {\em Proceedings of the IEEE/CVF Conference on Computer Vision and Pattern Recognition}, pages 22500--22510, 2023.

\bibitem{ruiz2023hyperdreambooth}
Nataniel Ruiz, Yuanzhen Li, Varun Jampani, Wei Wei, Tingbo Hou, Yael Pritch, Neal Wadhwa, Michael Rubinstein, and Kfir Aberman.
\newblock Hyperdreambooth: Hypernetworks for fast personalization of text-to-image models.
\newblock {\em arXiv preprint arXiv:2307.06949}, 2023.

\bibitem{saharia2022photorealistic}
Chitwan Saharia, William Chan, Saurabh Saxena, Lala Li, Jay Whang, Emily~L Denton, Kamyar Ghasemipour, Raphael Gontijo~Lopes, Burcu Karagol~Ayan, Tim Salimans, et~al.
\newblock Photorealistic text-to-image diffusion models with deep language understanding.
\newblock {\em Advances in Neural Information Processing Systems}, 35:36479--36494, 2022.

\bibitem{schuhmann2021laion}
Christoph Schuhmann, Richard Vencu, Romain Beaumont, Robert Kaczmarczyk, Clayton Mullis, Aarush Katta, Theo Coombes, Jenia Jitsev, and Aran Komatsuzaki.
\newblock Laion-400m: Open dataset of clip-filtered 400 million image-text pairs.
\newblock {\em arXiv preprint arXiv:2111.02114}, 2021.

\bibitem{shi2023instantbooth}
Jing Shi, Wei Xiong, Zhe Lin, and Hyun~Joon Jung.
\newblock Instantbooth: Personalized text-to-image generation without test-time finetuning.
\newblock {\em arXiv preprint arXiv:2304.03411}, 2023.

\bibitem{sohn2023styledrop}
Kihyuk Sohn, Nataniel Ruiz, Kimin Lee, Daniel~Castro Chin, Irina Blok, Huiwen Chang, Jarred Barber, Lu Jiang, Glenn Entis, Yuanzhen Li, et~al.
\newblock Styledrop: Text-to-image generation in any style.
\newblock {\em arXiv preprint arXiv:2306.00983}, 2023.

\bibitem{song2020denoising}
Jiaming Song, Chenlin Meng, and Stefano Ermon.
\newblock Denoising diffusion implicit models.
\newblock {\em arXiv preprint arXiv:2010.02502}, 2020.

\bibitem{tewel2023key}
Yoad Tewel, Rinon Gal, Gal Chechik, and Yuval Atzmon.
\newblock Key-locked rank one editing for text-to-image personalization.
\newblock In {\em ACM SIGGRAPH 2023 Conference Proceedings}, pages 1--11, 2023.

\bibitem{voynov2023p}
Andrey Voynov, Qinghao Chu, Daniel Cohen-Or, and Kfir Aberman.
\newblock $ p+ $: Extended textual conditioning in text-to-image generation.
\newblock {\em arXiv preprint arXiv:2303.09522}, 2023.

\bibitem{wang2023reprompt}
Yunlong Wang, Shuyuan Shen, and Brian~Y Lim.
\newblock Reprompt: Automatic prompt editing to refine ai-generative art towards precise expressions.
\newblock In {\em Proceedings of the 2023 CHI Conference on Human Factors in Computing Systems}, pages 1--29, 2023.

\end{thebibliography}
}

\newpage
\clearpage
\appendix

\section{Formatted prompt generation} \label{sec:formatted_prompt_generation}

For non-live objects, we use the following prompts to generate words describing shape, color, textures, and background in ChatGPT:

\begin{itemize}
    \item shape: give me 100 adjective words describing the shape of an object
    \item color: give me 100 adjective words describing the color of an object
    \item texture: give me 100 adjective words describing the texture of an object
    \item background: give me 500 phrases that describe the background, such as ``on the table", as diverse as possible.
\end{itemize}

After removing duplicated ones, there are 85 shapes, 93 colors, 96 textures, and 455 backgrounds.

For live objects, we use the following prompts to generate words describing shape, color, textures, and motion in ChatGPT:

\begin{itemize}
    \item body: give me 100 adjective words describing the body of an animal
    \item skin: give me 100 adjective words describing the skin or fur of an animal
    \item emotion: give me 100 adjective words describing the emotion of an animal
    \item motion: give me 1000 different short concise sentences that contains a special token ``\$concept" which stands for a specific animal, which can be a dog, a cat or a human. For example: ``a \$concept sitting in a temple", ``a \$concept walking in a supermarket". Keep ``a \$concept" in the sentences.
\end{itemize}

After removing duplicated ones, there are 89 bodies, 86 skins/furs, 75 emotions, and 744 motions. For humans, we replace the word "animal" above with "person".

We use 

\begin{itemize}
    \item style: give me 100 image style descriptions, such as ``a photo of", and ``a painting of".
\end{itemize}

After removing duplicated ones, there are 99 styles left.

\section{Better Category Naming}~\label{sec:better_naming}
\begin{figure}
% \vspace{-4ex}
\begin{center}
  \includegraphics[width=0.3\textwidth]{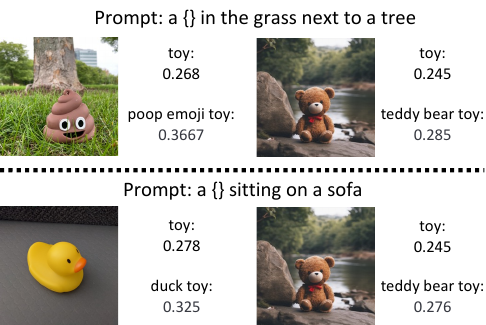}
\end{center}
\vspace{-2ex}
   \caption{\small \textbf{CLIP-T score on different subject names}. We randomly generate an image with the prompt ``a teddy bear toy sitting beside a river''. Subsequently, we evaluate prompts ``a \{\} in the grass next to a tree'' and ``a \{\} sitting on a sofa''. When employing the vague prompt ``toy'', both the ground truth and the mismatched example are classified as mismatches (\textless0.3). In contrast, when using specific class names, the ground truth is categorized as a match (\textgreater0.3), while the mismatched examples remain classified as mismatches.}
\label{fig:clip_metric}
\end{figure}

We show the name change in Table~\ref{tab:name_change}. As shown in Figure~\ref{fig:clip_metric}, there is a notable discrepancy between the utilization of vague class names, such as ``toy'', and more specific object names, such as ``duck toy'', on the ground truth images. Notably, the CLIP-T score appears to be significantly influenced by the nomenclature chosen for the object, and thereby potentially undermines its accuracy as an indicator of text-image alignment. To delve deeper into this matter, we calculate the CLIP-T score on the original images and the manually added prompts. Table~\ref{tab:clip_matrics} presents that, when using vague class names, the CLIP-T score for ground truth text-image pairs falls even below the conventional threshold of 0.3, which is typically considered as the threshold for assessing text-image pair compatibility~\cite{schuhmann2021laion}. To rectify this issue, we replace vague class names with highly specific object names, which results in a substantial improvement in the CLIP-T score for ground truth text-image pairs. 
Additional details regarding the modified nomenclature list can be found in Appendix~\ref{sec:better_naming}.

\begin{table}[!thb]
\centering
\small
% \resizebox{0.98\linewidth}{!}{%
\begin{tabular}{@{\;}lcc@{\;}}
\toprule
subject name & original class & modified class\\
\midrule
bear\_plushie & stuffed animal &  bear plushie \\
berry\_bowl & bowl  & berry bowl \\
can & can & drink can \\
clock & clock & alarm clock \\
duck\_toy & toy & duck toy \\
grey\_sloth\_plushie & stuffed animal & sloth plushie \\
monster\_toy & toy & monster toy \\
poop\_emoji & toy & poop emoji toy \\
rc\_car & toy & racing car toy \\
red\_cartoon & cartoon & 2d cartoon devil \\
robot\_toy & toy & robot toy \\
wolf\_plushie & stuffed animal & wolf plushie \\
\bottomrule
\end{tabular}
% }
% \vspace{1ex}
\caption{\small
\textbf{Name Change}. We change the name for a more reasonable CLIP-T metric and better performance.
}
% \vspace{-4ex}
\label{tab:name_change}
\end{table}

\section{Details of User Study}~\label{sec:user_study}
\begin{figure*}[h]
\centering
\includegraphics[width=0.95\textwidth]{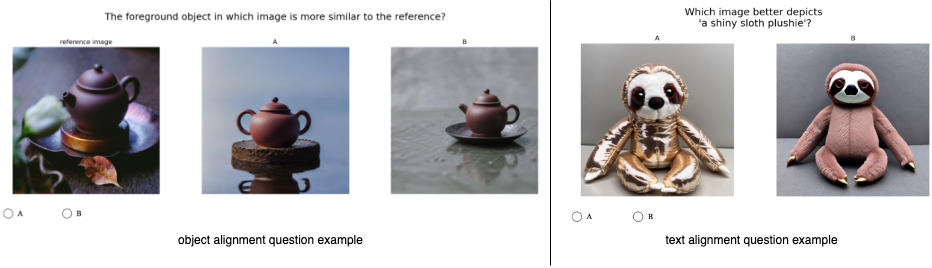}
\caption{An example of the user study interface. The left is an example for question of object alignment, and the right is an example for question of text alignment. Each user was asked to answer 30 questions
about subject alignment (left) and 30 questions about text alignment (right). The examples in the questions are randomly sampled from a large pool.}
\label{fig:user_study_interface}
\end{figure*}

We randomly sampled and paired 300 comparisons of ours(SD) versus DreamBooth, half of which is for the subject alignment and the other half for the text alignment. For subject alignment, we randomly sampled a ground truth image and asked "The foreground object in which image is more similar to the reference?". For text alignment, we asked "Which image better depicts \{\}?", where \{\} is replaced by the prompt. We equally divided the questions into 10 groups. Each person randomly received one group. We did the same for ours(SDXL) versus ours(SD). We provide an example of our interface in Figure~\ref{fig:user_study_interface}.

\section{Extension: Using BLIP to Generate Captions}
\begin{figure*}[t]
\begin{center}
  \includegraphics[width=0.8\textwidth]{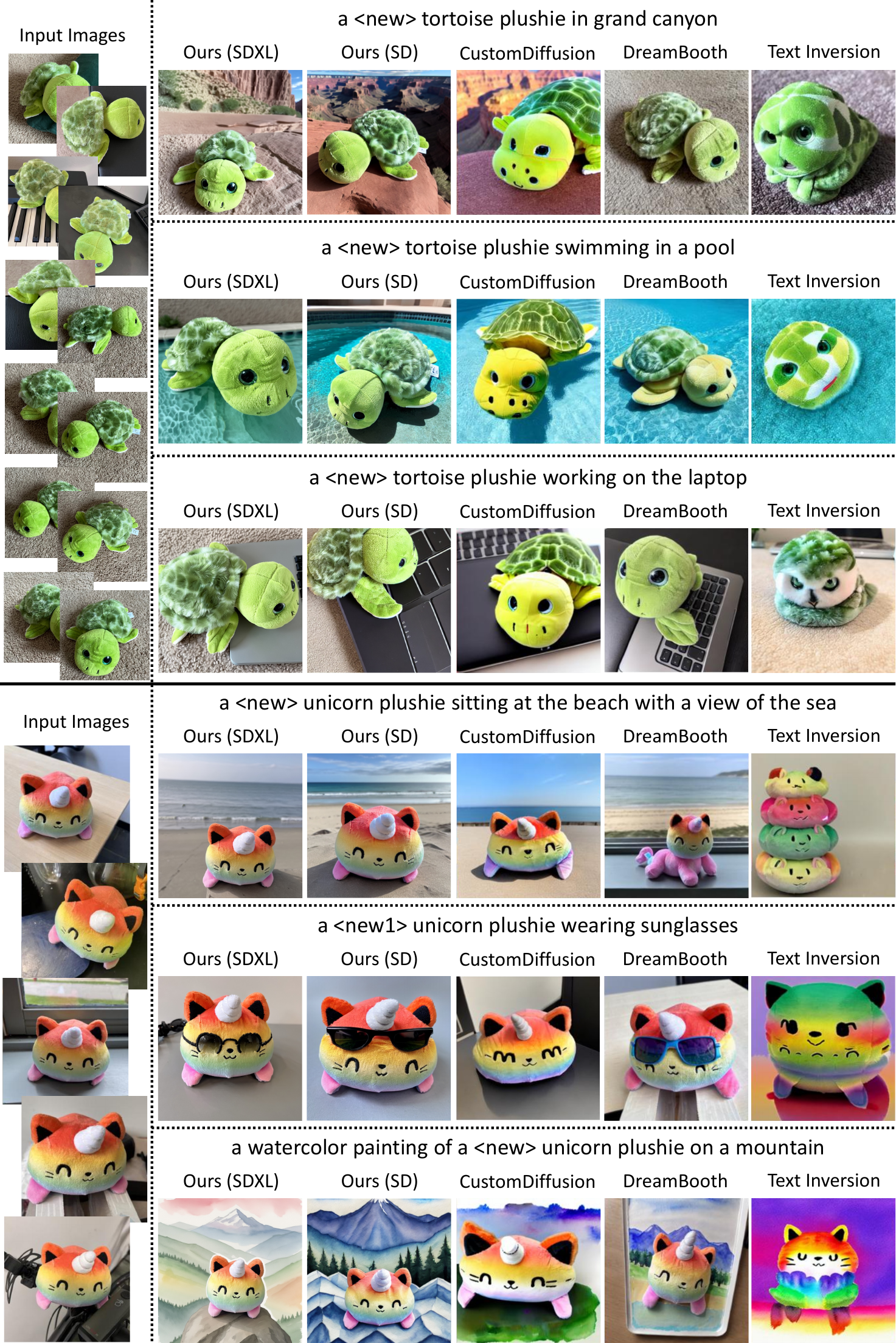}
\end{center}
   \caption{More Comparison.}
\label{fig:more_comp2}
\end{figure*}

We tried to use BLIP~\cite{li2022blip} to generate more personalized captions for the training example images. BLIP outputs a caption for the input image and can be conditioned on the format. 
We condition BLIP 
%on the prompt "a [subject]" 
so that it generates prompts that start with "a [subject]". 
%Then task ChatGPT to insert the special token before the "[subject]". 
For instance for the "tortoise plushie" image BLIP generates
\begin{itemize}
    \item a tortoise plushie on a pillow
    \item a tortoise plushie
    \item a tortoise plushie sitting on a piano keyboard
    \item a tortoise plushie on a desk
    \item ...
\end{itemize}
To unify the prompt format, we task ChatGPT to\textit{`Change the following sentence to the format "A $<$new$>$ tortoise plushie blablabla". The "$<$new$>$" is a special token that needs to be inserted
%as "$<$new$>$" itself 
before tortoise plushie.'} The result are the following prompts
\begin{itemize}
    \item a $<$new$>$ tortoise plushie on a pillow
    \item a $<$new$>$ tortoise plushie
    \item a $<$new$>$ tortoise plushie sitting on a piano keyboard
    \item a $<$new$>$ tortoise plushie on a desk
    \item ...
\end{itemize}
The results of this "tortoise plushie" dataset is shown in Figure~\ref{fig:more_comp2}. 
With this addition of using BLIP, it alleviated writing the prompt examples manually, i.e., it replaced the manual steps in Section~\ref{sec:method}.

\section{Implementation Details}~\label{sec:implementation_details}
We opt for the identifier word ``olis'' instead of the more commonly used ``sks''. This choice is based on the fact that ``olis'' corresponds to the least frequently utilized token in the model's vocabulary~\cite{special_token}. 
Each training batch contains one example from training set and one example from regularization set.
For SD, we fine-tune the entire model with a learning rate of 2e-6 and perform inference using 200 steps of DDIM~\cite{song2020denoising}.
For SDXL, which has a larger model size, we employ a LoRA with a rank of 32 for both the text encoders and UNet. We also train the text embeddings. We set learning rate to 1e-4. We use 50 steps of DDIM for inference.
We show the best number of iterations in Table~\ref{tab:best_iteration}. For simplicity, we use 4000 and 8000 iterations for SD and SDXL, respectively.
\begin{table}[!thb]
\centering
\small
% \resizebox{0.98\linewidth}{!}{%
\begin{tabular}{@{\;}lcc@{\;}}
\toprule
subject name & best \#iterations on SD & best \#iterations on SDXL\\
\midrule
backpack & 6000-8000  & 8000-10000 \\
backpack\_dog & 2000-3000 & 4000-6000 \\
bear\_plushie & 2000-4000 & 4000-6000 \\
berry\_bowl & 6000-8000 & 8000-10000\\
can & 6000-8000 & 8000-10000 \\
candle & 4000-6000 & 8000-10000\\
cat & 1000-3000 & 1000-3000\\
cat2 & 6000-8000 & 8000-10000\\
clock & 6000-8000 & 8000-10000\\
colorful\_sneaker & 4000-6000 & 6000-8000 \\
dog & 1000-3000 & 1000-3000\\
dog2 & 2000-4000 & 4000-6000\\
dog3 & 2000-4000 & 8000-10000\\
dog5 & 3000-4000 & 6000-8000 \\
dog6 & 3000-4000 & 6000-8000\\
dog7 & 3000-4000 & 6000-8000 \\
dog8 & 1000-3000 & 1000-3000\\
duck\_toy & 3000-4000 & 3000-4000 \\
fancy\_boot & 3000-4000 & 6000-8000 \\
grey\_sloth\_plushie & 3000-4000 & 6000-8000 \\
monster\_toy & 3000-4000 & 8000-10000\\
pink\_sunglasses & 3000-4000 & 4000-6000 \\
poop\_emoji & 3000-4000 & 4000-6000 \\
rc\_car & 3000-4000 & 4000-6000 \\
red\_cartoon & 6000-8000 & 8000-10000\\
robot\_toy & 3000-4000& 6000-8000 \\
shiny\_sneaker & 3000-4000& 6000-8000\\
teapot & 6000-8000& 8000-10000 \\
vase & 6000-8000 & 8000-10000 \\
wolf\_plushie & 3000-4000 & 4000-6000 \\
\bottomrule
\end{tabular}
% }
% \vspace{1ex}
\caption{\small
\textbf{Best \#iterations of datasets in DreamBench.} The variation mainly comes from the diversity of the dataset itself.
}
% \vspace{-4ex}
\label{tab:best_iteration}
\end{table}

\section{More Results}~\label{sec:more_results}

\begin{figure*}
\begin{center}
  \includegraphics[width=1\textwidth]{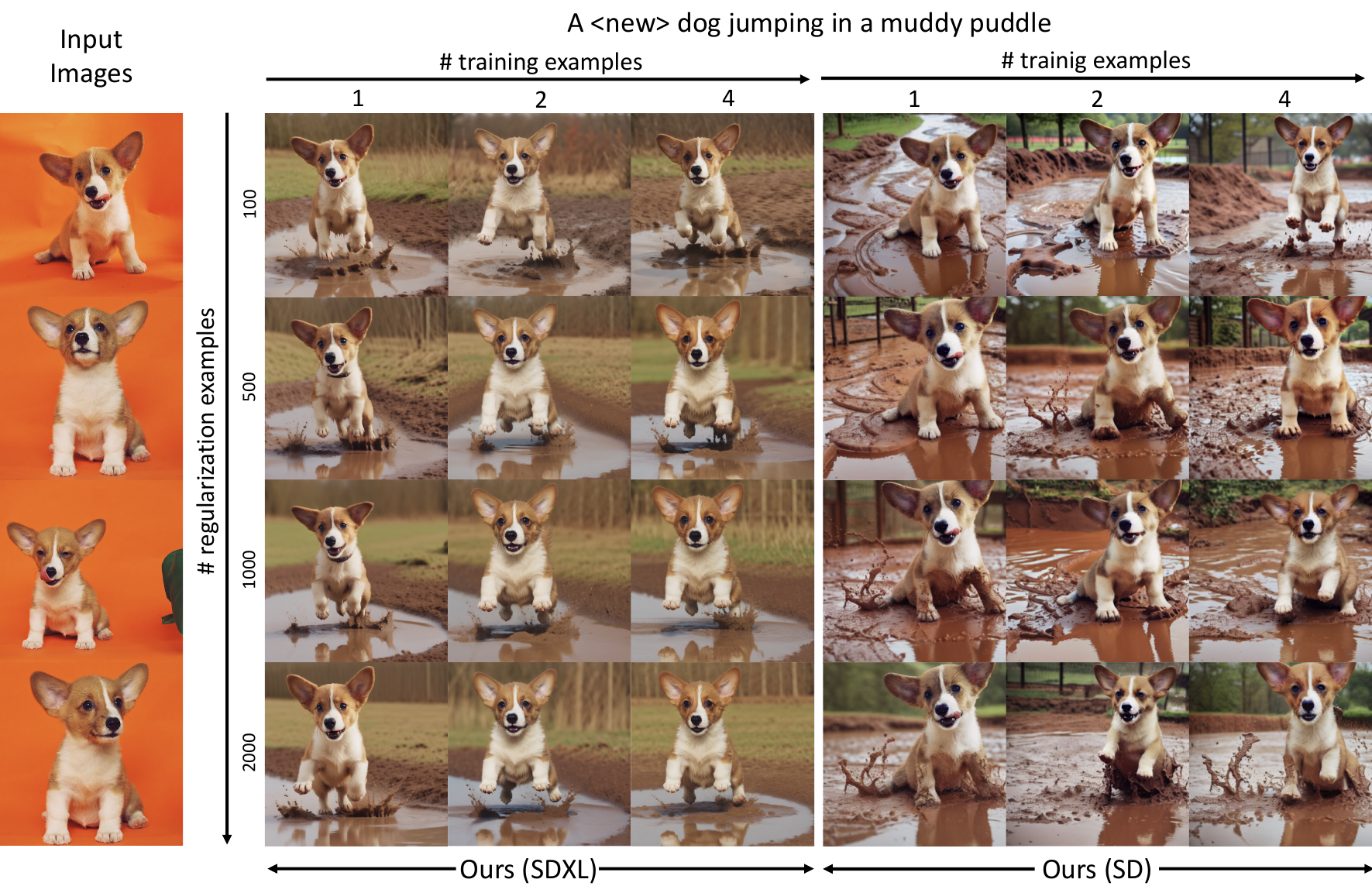}
\end{center}
   \caption{Ablation Tests on number of training examples and regularization dataset size. Even only a very small regularization dataset is given (\~100 examples), our method still effectively prevent overfitting and preserves the identity.}
\label{fig:ablation}
\end{figure*}

\begin{figure*}
\begin{center}
  \includegraphics[width=\textwidth]{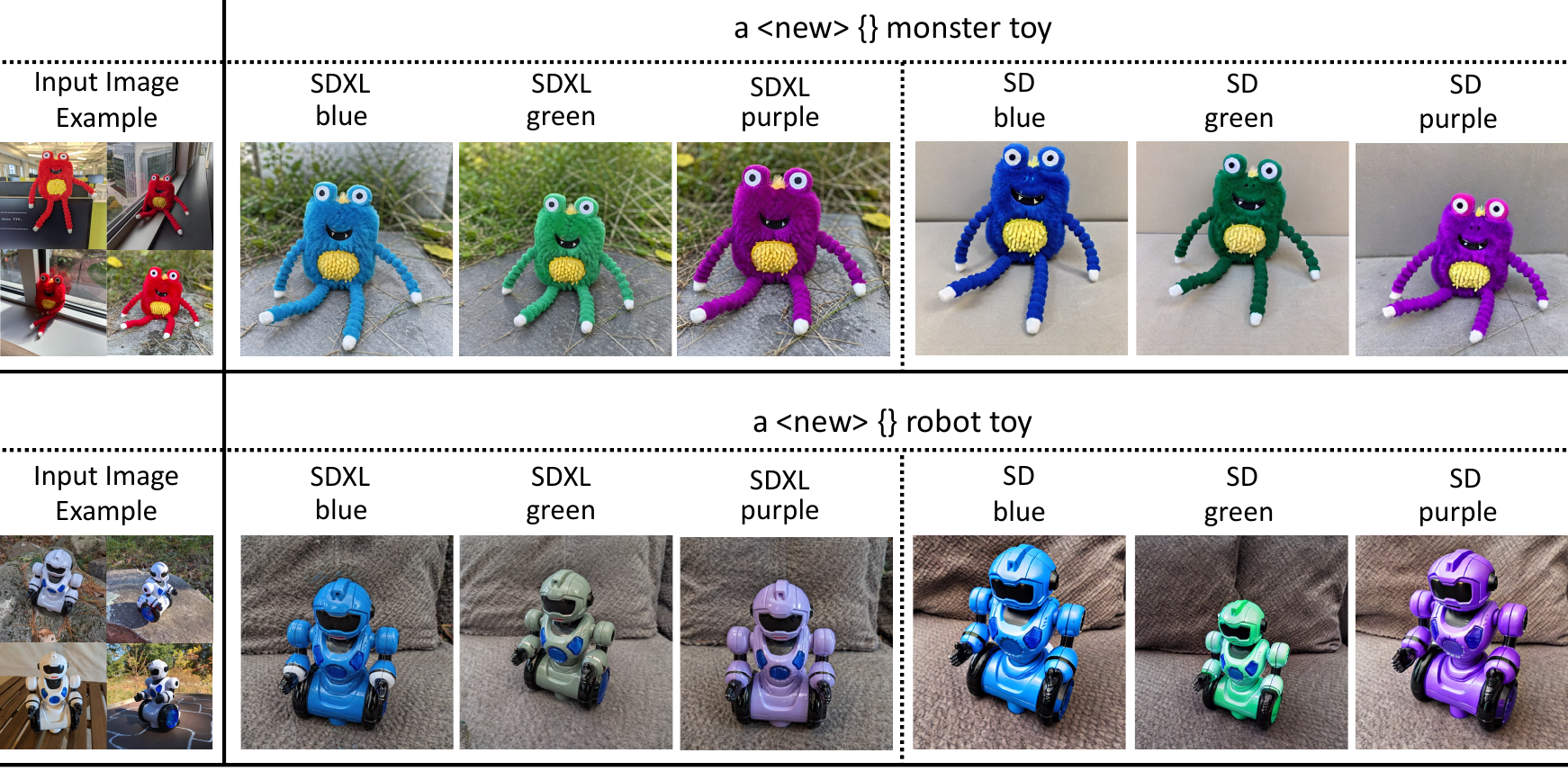}
\end{center}
\vspace{-0.2cm}
   \caption{\textbf{Color Modification}. Our method can alter the color of the subject. It is important to mention that when modifying the color, using ``a \textless new\textgreater [color] [class noun]'' is more effective than ``a [color] \textless new\textgreater [class noun]''.}
\label{fig:color_modification}
\end{figure*}

\begin{figure*}[t]
\begin{center}
  \includegraphics[width=0.9\textwidth]{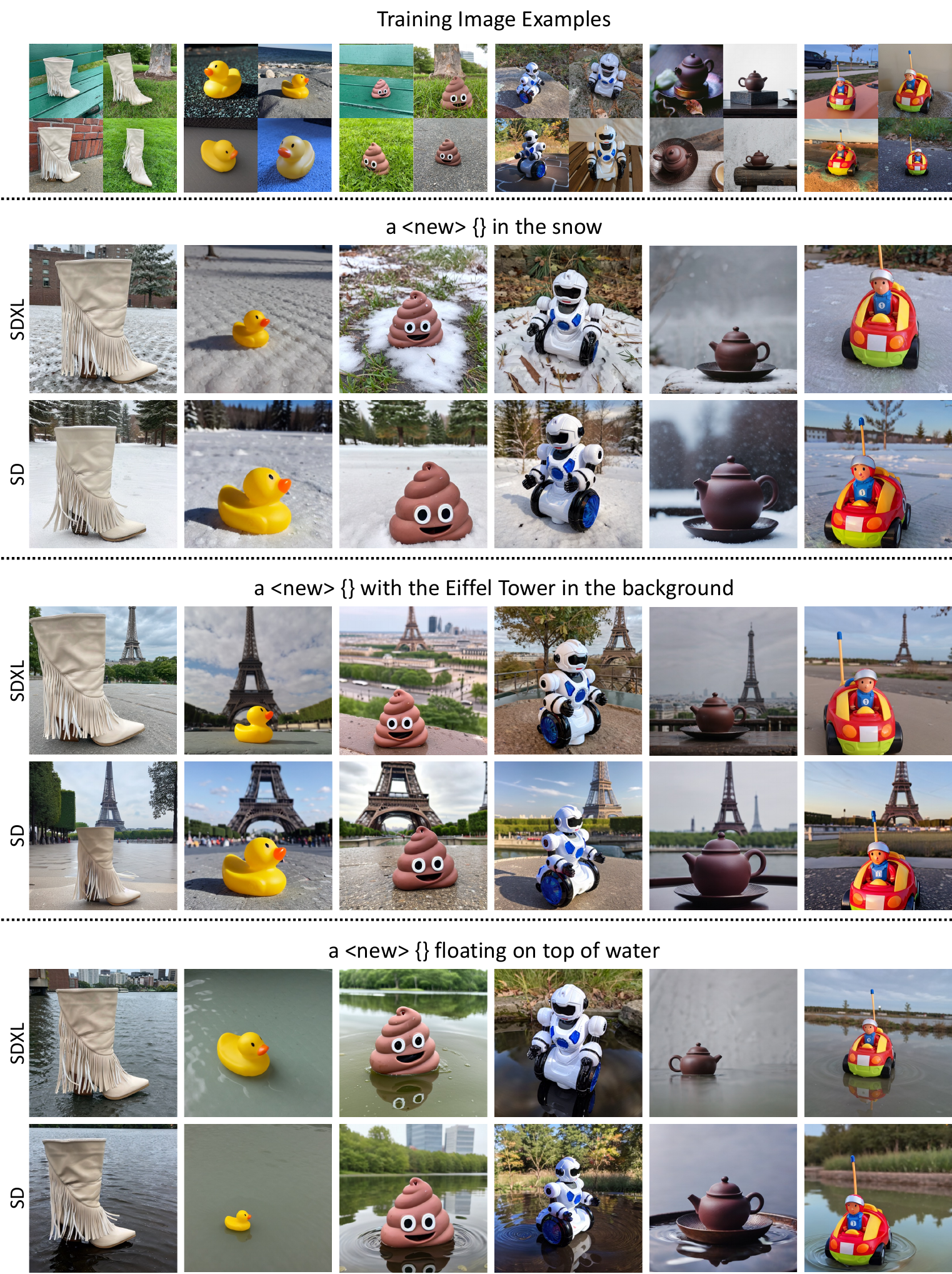}
\end{center}
   \caption{More Results on inanimate objects.}
\label{fig:more1}
\end{figure*}

\begin{figure}[t]
\begin{center}
  \includegraphics[width=0.9\textwidth]{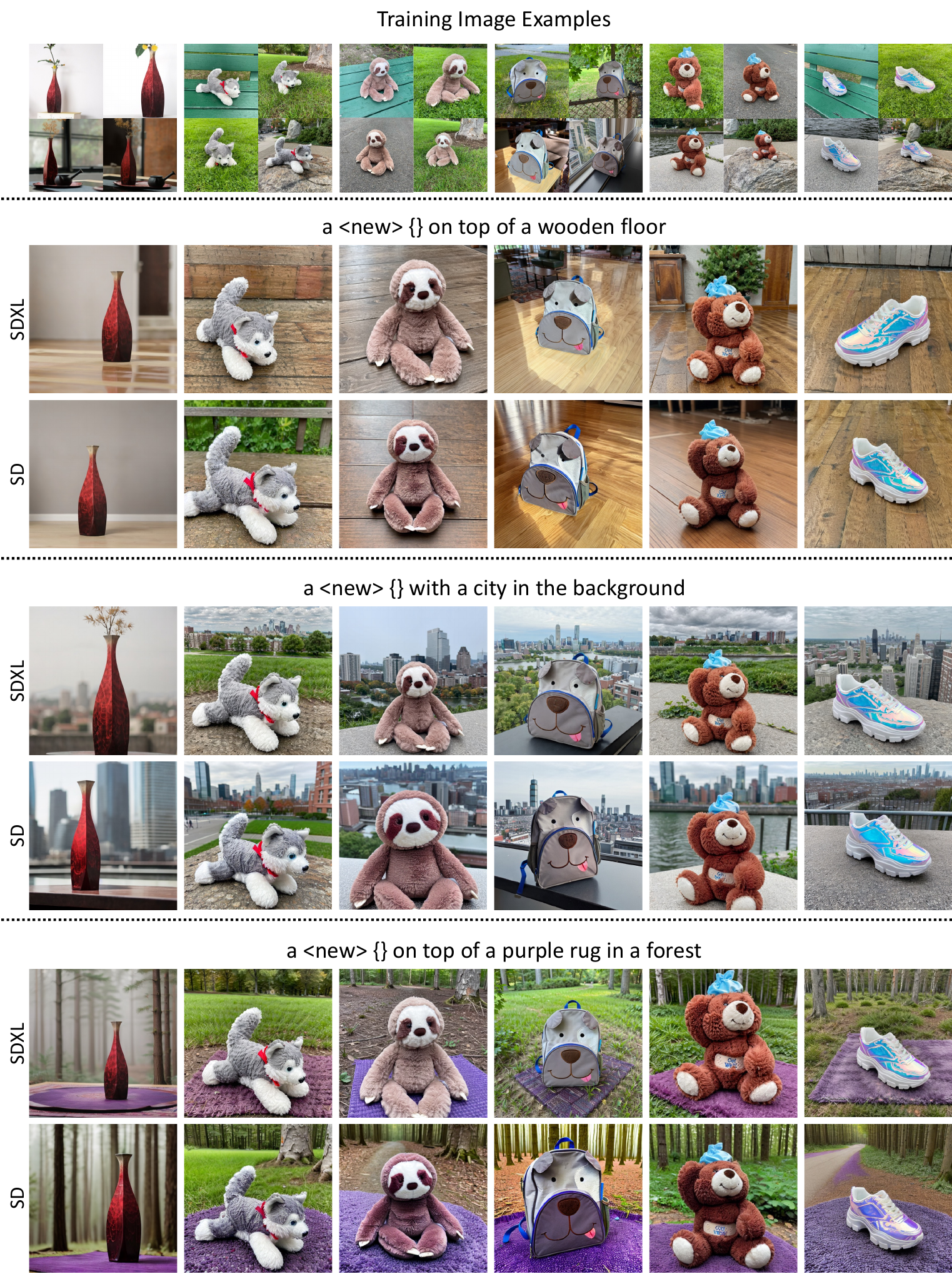}
\end{center}
   \caption{More Results on inanimate objects.}
\label{fig:more4}
\end{figure}

\begin{figure*}[t]
\begin{center}
  \includegraphics[width=0.9\textwidth]{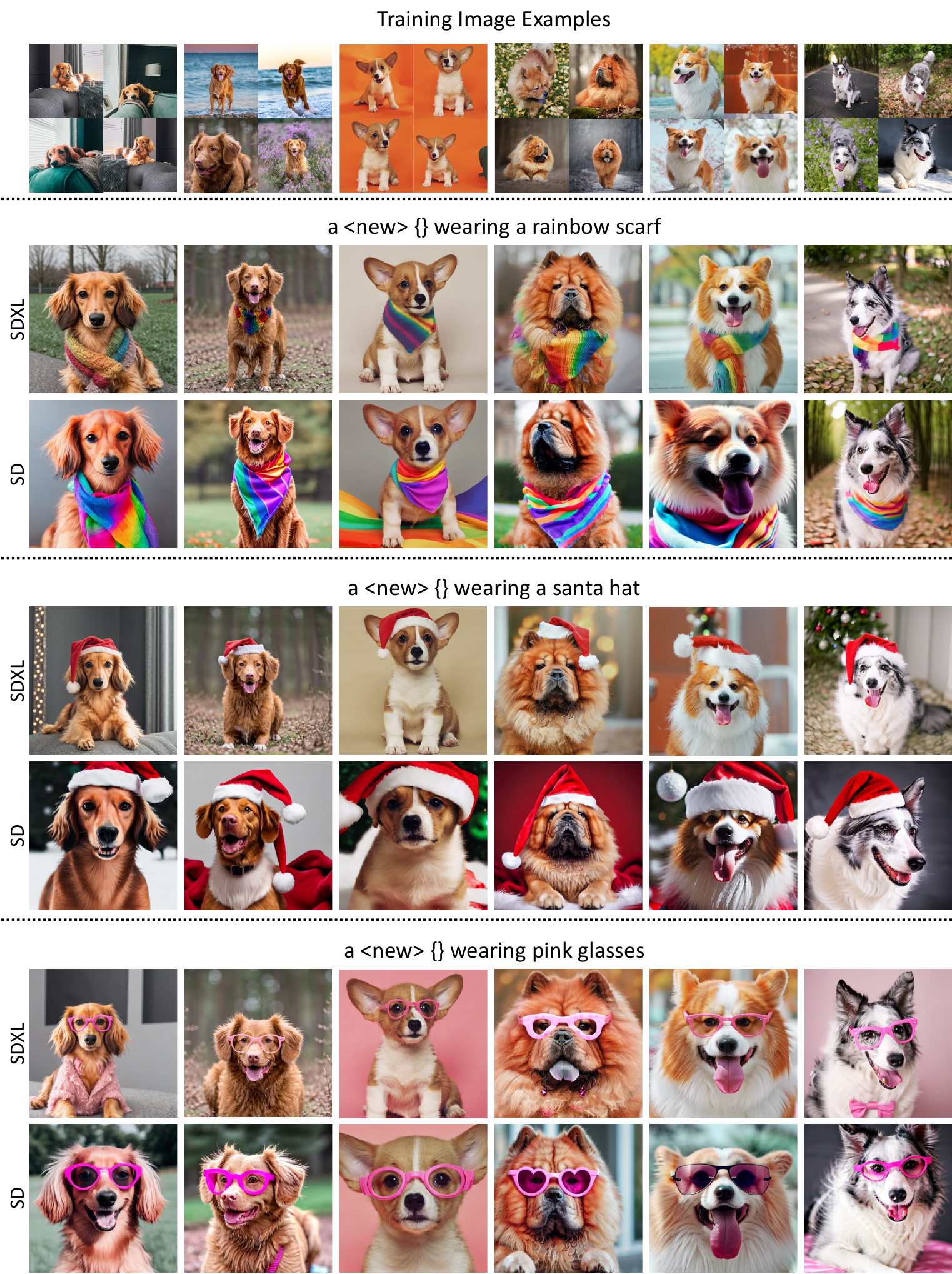}
\end{center}
   \caption{More Results on living entities.}
\label{fig:more2}
\end{figure*}

\begin{figure*}[t]
\begin{center}
  \includegraphics[width=0.9\textwidth]{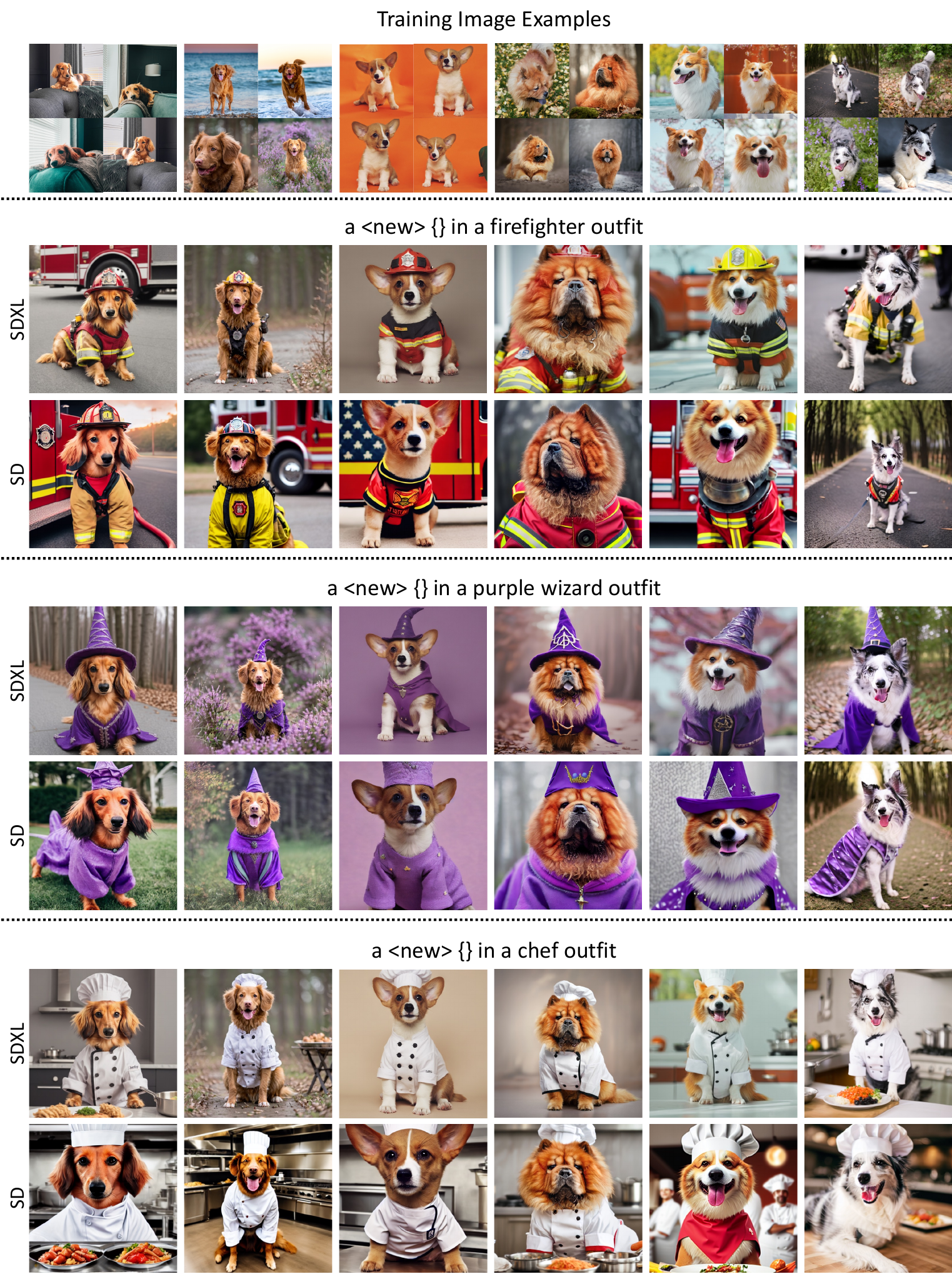}
\end{center}
   \caption{More Results on living entities.}
\label{fig:more3}
\end{figure*}

\begin{figure*}[t]
\begin{center}
  \includegraphics[width=0.8\textwidth]{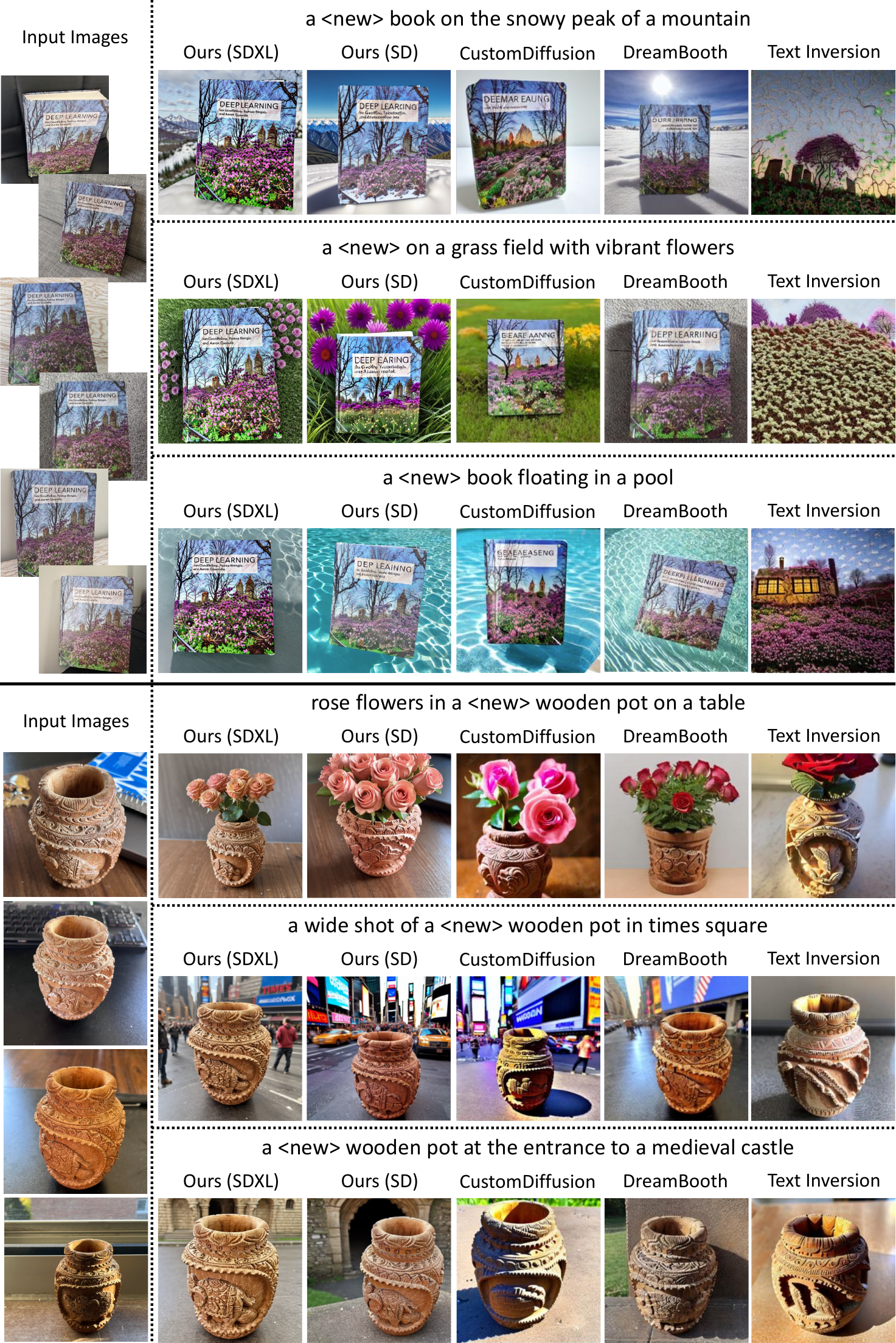}
\end{center}
   \caption{More Comparison.}
\label{fig:more_comp1}
\end{figure*}

\begin{figure*}[t]
\begin{center}
  \includegraphics[width=0.8\textwidth]{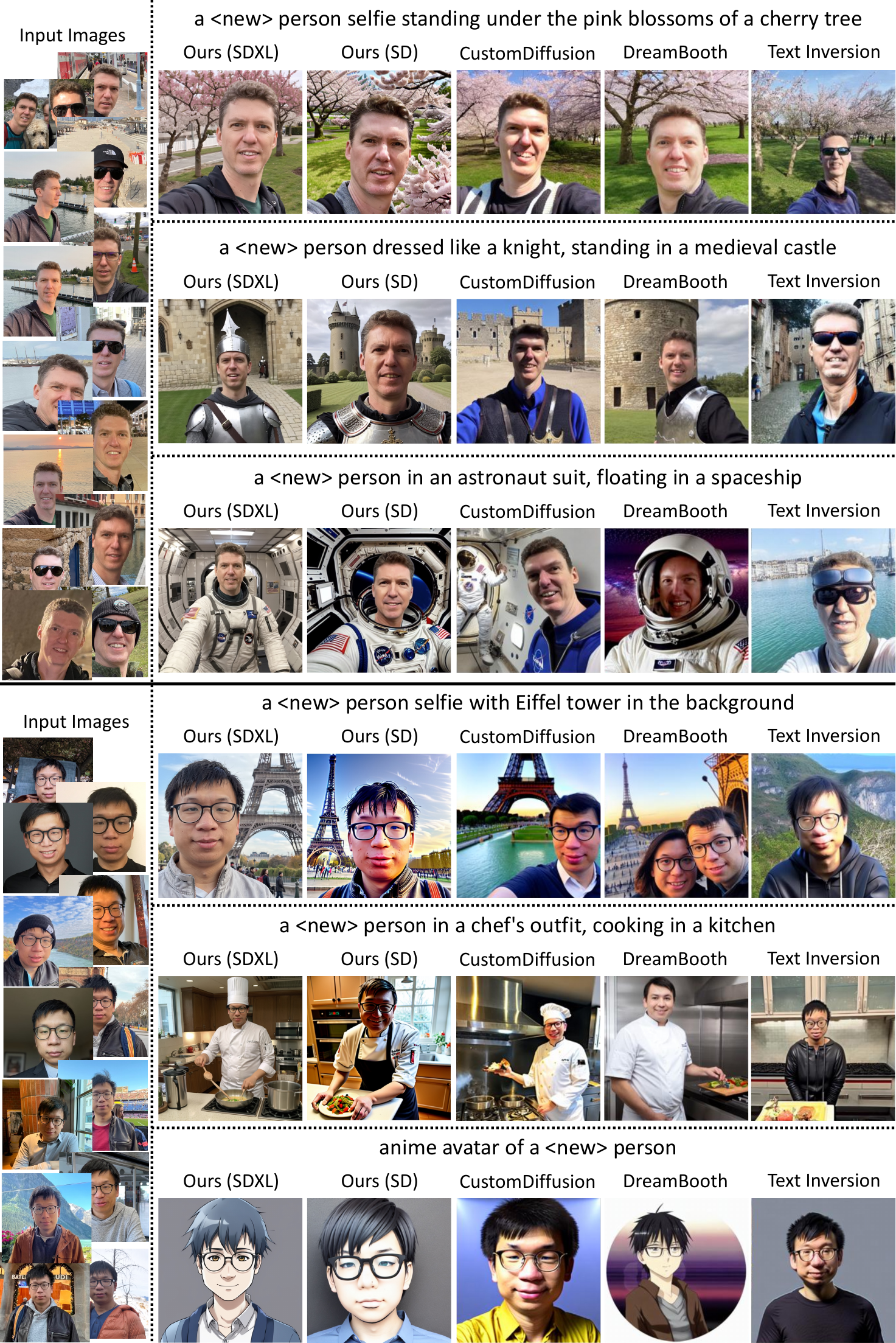}
\end{center}
   \caption{More Comparison.}
\label{fig:more_comp3}
\end{figure*}

\end{document}